\documentclass[11pt]{article}
\usepackage[preprint]{acl} 

\usepackage{times}
\usepackage{latexsym}
\usepackage[T1]{fontenc}
\usepackage[utf8]{inputenc}
\usepackage{microtype}
\usepackage{inconsolata}

\usepackage{amsmath,amssymb}
\usepackage{braket}
\usepackage{graphicx}
\DeclareGraphicsExtensions{.pdf,.png,.jpg}

\usepackage{booktabs}
\usepackage{tabularx}
\usepackage{multirow}
\usepackage{float}
\usepackage{xcolor}
\usepackage{soul}
\usepackage{algorithm}
\usepackage{algpseudocode}
\usepackage{tcolorbox}
\tcbuselibrary{breakable,skins}
\usepackage{adjustbox}
\usepackage{listings}

\hypersetup{
  colorlinks=true,
  linkcolor=blue,
  citecolor=blue,
  urlcolor=blue
}

\algrenewcommand\algorithmicindent{0.6em}
\algrenewcommand\alglinenumber[1]{\scriptsize #1}
\makeatletter
\def\ALG@linenumber@l#1{\makebox[2.2em][r]{\scriptsize #1}\hspace*{0.35em}}
\makeatother
\newcommand{\WrapState}[1]{\State \parbox[t]{\dimexpr\linewidth-2.55em\relax}{#1}}

\newtcolorbox{promptbox}[1][]{
  enhanced,
  breakable,                  
  colback=black!1!white,
  colframe=black!57!white,
  boxrule=0.3pt,
  top=1mm, bottom=1mm,
  left=1mm, right=1mm,
  before skip=6pt plus 2pt minus 1pt,
  after skip=6pt plus 2pt minus 1pt,
  width=\columnwidth,         
  #1
}

\begin{document}

\title{PluriHopRAG: Exhaustive, Recall-Sensitive QA Through Corpus-Specific Document Structure Learning}

\author{
  Mykolas Sveistrys\thanks{Corresponding author} \\
  Turbit Systems GmbH \\
  \texttt{m.sveistrys@turbit.de}
  \And
  Richard Kunert \\
  Turbit Systems GmbH \\
  \texttt{r.kunert@turbit.de}
}

\maketitle

\begin{abstract}
     Retrieval-Augmented Generation (RAG) has been used in question answering (QA) systems to improve performance when relevant information is in one (single-hop) or multiple (multi-hop) passages. However, many real life scenarios (e.g. dealing with financial, legal, medical reports) require checking all documents for relevant information without a clear stopping condition. We term these pluri-hop questions, and formalize them by 3 conditions - recall sensitivity, exhaustiveness, and exactness. To study this setting, we introduce PluriHopWIND, a multilingual diagnostic benchmark of 48 pluri-hop questions over 191 real wind-industry reports, with high repetitiveness to reflect the challenge of distractors in real-world datasets. Naive, graph-based, and multimodal RAG methods only reach up to 40\% statement-wise F1 on PluriHopWIND. Motivated by this, we propose PluriHopRAG, which learns from synthetic examples to decompose queries according to corpus-specific document structure, and employs a cross-encoder filter at the document level to minimize costly LLM reasoning. We test PluriHopRAG on PluriHopWIND and the Loong benchmark built on financial, legal and scientific reports. On PluriHopWIND, our method shows 18-52\% F1 score improvement across base LLMs, while on Loong, we show 33\% improvement over long-context reasoning and 52\% improvement over naive RAG.
\end{abstract}

\begin{figure*}[!t]
    \centering
    \includegraphics[width=0.95\linewidth]{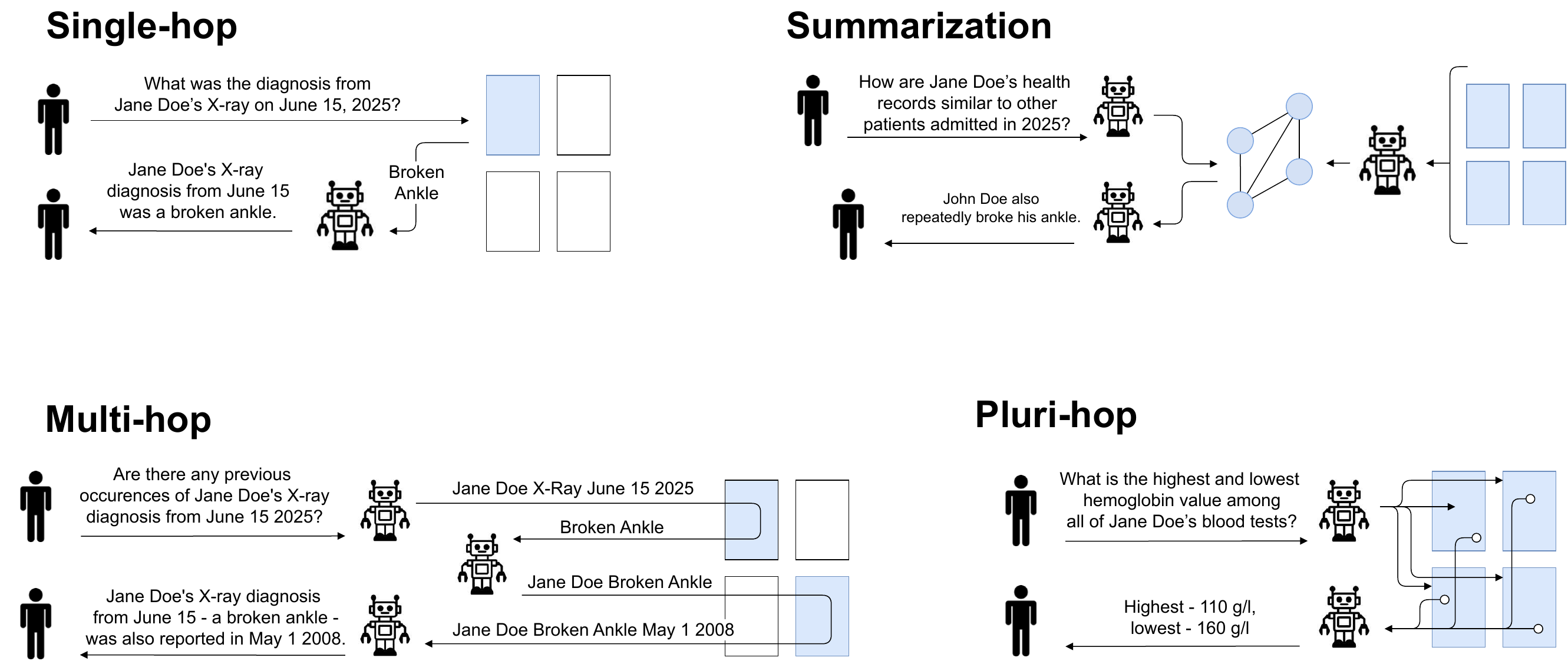}
    \caption{Common types of questions RAG systems are used for.}
    \label{fig:question_taxonomy}
\end{figure*}

\section{Introduction}

The rise of Large Language Models (LLMs) \cite{gpt3} has enabled rapid progress in question answering (QA) systems, by incorporating LLMs into a QA framework called Retrieval Augmented Generation (RAG) \cite{rag-review-2023}. The strength of RAG lies in combining information retrieval techniques with an LLM's ability to synthesize chunks of evidence into a human-like answer. Over time, the scope of RAG has expanded, allowing it to tackle increasingly complex types of questions.

Early RAG systems \cite{RAG-original-paper} were best suited for \textbf{single-hop} questions - questions with only one or several relevant passages - because they simply searched for passages that are semantically similar to the original question. Iterative improvements have enabled progress on \textbf{multi-hop} questions where one piece of evidence informs the retrieval of the next; they are often addressed through iterative, agentic, and planning-based approaches \cite{ircot, iter-retgen, self-rag}. A parallel line of work has tackled global \textbf{summarization-style} questions by using knowledge-graph-based RAG approaches \cite{gnn-rag, grag, graphrag} that leverage structured entity-relationship representations. These question types are illustrated with examples in Figure \ref{fig:question_taxonomy}.

In contrast, there has been considerably less progress on a fourth category: questions that require aggregating data across all documents in the knowledge base (see Figure \ref{fig:question_taxonomy}). For example, in the context of medical records: "What is the highest and lowest hemoglobin value among all of Jane Doe's blood tests?". Unlike conventional multi-hop queries, these problems lack a natural stopping condition - retrieval cannot halt after a handful of documents because every record may change the answer, and unlike summarization-style questions, they have an exact answer.

In this work, we focus on precisely these questions and coin them \textbf{pluri-hop} questions. They are defined by three conditions:
\begin{enumerate}
    \item Recall sensitivity: Omitting even a single relevant passage leads to an incorrect answer.
    \item Exhaustiveness: It is impossible to infer from the retrieved context whether the evidence set is complete; in principle, all documents must be checked.
    \item Exactness: There is only one best answer. All other answers are either incomplete or contain superfluous and/or incorrect information.
\end{enumerate}

These conditions imply that a viable approach to pluri-hop QA must go beyond existing paradigms which are typically based on "top-k" retrieval. Instead of selectively focusing on a small subset of passages, the system must be designed to check all documents efficiently, while filtering irrelevant material early to maintain feasibility.

Despite a lack of targeted investigations, pluri-hop questions are widespread, especially when handling recurring report data - medical records, financial reports, compliance reports, etc. - see Table \ref{tab:pluri-hop-examples} for a list of examples. Such data poses a challenge to RAG systems due to a large presence of distractor documents - documents that, given a question, are irrelevant but semantically similar to relevant documents, thus "distracting" the RAG system. Therefore, in this work, we seek to answer the following question: \textbf{How does one answer pluri-hop questions about highly repetitive data (such as data from recurring reports) in a scalable way?}

To highlight the difficulties of answering pluri-hop questions, we introduce \textbf{PluriHopWIND}, a diagnostic multilingual dataset of 48 questions constructed from 191 real-world wind industry technical reports in German and English. Crucially, many benchmark questions require consulting evidence spanning more than the context window of state-of-the-art LLMs. The dataset also emphasizes distractor density, with large amounts of semantically similar but irrelevant material, closely mirroring practical QA challenges in recurring report corpora. \textbf{We show that current approaches struggle to answer pluri-hop questions, reaching at most 40\% statement-wise F1 score.}

Motivated by this, we propose \textbf{PluriHopRAG}, built specifically for pluri-hop questions by following two design principles:
\begin{enumerate}
    \item \textbf{Structure-aware query decomposition}: the system must learn how information is distributed across documents, then split queries into document-level subquestions accordingly, by using a query decomposer fine-tuned on a small number of synthetic examples.
    \item \textbf{Cheap document-level filtering}: since all documents must be checked, we employ cross-encoder filtering to discard irrelevant documents after chunk retrieval but before expensive LLM reasoning.
\end{enumerate}

We compare our approach to a baseline RAG approach, RAG based on knowledge graphs (GraphRAG \cite{graphrag}), and vision models (VisdomRAG \cite{visdomrag}). \textbf{PluriHopRAG outperforms competing approaches by 18-52\% on benchmark F1 score across base LLMs.}

To demonstrate the general utility of PluriHopRAG, we use the Loong QA dataset \cite{loong} containing research papers, financial reports, and legal documents. Loong QA emphasizes aggregation across many passages of text and high recall sensitivity. \textbf{PluriHopRAG outperforms long-context QA by 33\% and the previously top-scoring model \cite{structrag} by 14\%.}

Taken together, our findings suggest that pluri-hop QA is insufficiently addressed by prominent RAG approaches. Despite its modest size, the PluriHopWIND dataset exposes the limitations of current QA systems on repetitive, distractor-rich corpora, while PluriHopRAG’s gains highlight the value of exhaustive retrieval with early filtering as a powerful alternative to top-k methods.

\begin{table*}[!t]
\centering
\begin{tabular}{p{3cm}p{4cm}p{8cm}}\toprule
\textbf{Sector} & \textbf{Document Type} & \textbf{Typical Question (pluri-hop)} \\ \hline

Healthcare & Lab results &
Across 2022–2024, what are Jane Doe’s lowest and highest eGFR values, with dates? \\

Education & Student progress report &
Which students failed two or more terms between Fall 2022 and Spring 2025? \\

Energy \& Utilities & Turbine inspection report &
In windpark W03 (2022–2024), which turbine has the most gearbox-wear reports (moderate+)? \\

Retail \& Supply Chain & Supplier compliance report &
Which suppliers had quality-check failures in 3+ separate audits (2022–2024)? \\

Legal \& Contracts & Compliance audit &
For Contract C-17 (2019–2025), which clauses were ever marked non-compliant? \\ \hline
\end{tabular}
\caption{Examples of pluri-hop questions about recurring-report corpora from various fields.}
\label{tab:pluri-hop-examples}
\end{table*}

\section{Related Work}

\textbf{Methods.}
Iterative approaches (IRCoT \cite{ircot}, Iter-RetGen \cite{iter-retgen}, Self-RAG \cite{self-rag}) break questions into sub-queries but underperform when scanning the entire corpus is required. Graph-based approaches (GRAG \cite{grag}, GNN-RAG \cite{gnn-rag}, GraphRAG \cite{graphrag}) enable multi-hop reasoning through knowledge graphs but often lose fine-grained details in raw text. Multi-modal methods like VisdomRAG \cite{visdomrag} incorporate visual layout cues but do not address scalability for large, repetitive corpora.

\textbf{Benchmarks.}
Multi-hop benchmarks such as HotpotQA \cite{hotpotqa}, 2WikiMultiHopQA \cite{2wikimultihopqa} and MultiHopRAG \cite{multihoprag} evaluate systems on linking evidence from multiple passages. However, these questions have clear stopping conditions - given the retrieved context, one can determine whether it is sufficient, violating the exhaustiveness criterion of pluri-hop questions. Summarization-oriented benchmarks such as NarrativeQA \cite{narrativeqa} require models to condense long narratives into high-level answers, violating the exactness and recall sensitivity criteria.

MoNaCo \cite{monaco} uses Wikipedia, confounding retrieval evaluation with pretraining knowledge - the authors found that adding retrieval actually degraded performance compared to an LLM-only baseline.

The Loong benchmark \cite{loong} is conceptually similar to our work as it requires aggregating many passages of text and focusses on recall sensitivity. However, the benchmark was created to test long-context reasoning, not scalable retrieval, so the context for each document fits within an LLM's context window. 

\textbf{In summary,} existing methods focus on (i) RAG with clear stopping conditions, (ii) RAG for summarization questions, or (iii) passing the full corpus to an LLM. None address pluri-hop questions requiring exhaustive, recall-sensitive aggregation across large, repetitive, distractor-heavy corpora in a scalable way. This motivates our introduction of the PluriHopWIND dataset and the PluriHopRAG model.

\section{Dataset}

\subsection{QA Generation}

PluriHopWIND consists of 48 English questions from 191 technical reports in German or English. The reports originate from the wind industry covering oil laboratory analyses, turbine inspections, and service activities. Each document has been anonymized (blacked out personally identifiable information) and pseudonimized (renamed turbines and windparks, some dates shifted). The documents vary highly in length (1-50 pages) and structure. However, almost all documents combine multiple visually-rich elements, like complex tables, diagrams, images and pictograms, while also containing whole paragraphs of text; see Figure \ref{fig:typical_report} for an example page from an oil analysis report.

We generate pluri-hop questions via a two-step process. First, we manually create 2-7 single-hop question-answer pairs per document, designed to extract information from visually-rich elements (tables, diagrams) and reflect each report category's function. Second, an LLM aggregates these into pluri-hop questions\footnote{We also instruct the LLM that each reference to a document should be quoted with its filename. This is used when calculating the efficacy of the cross-encoder filter.} that should:

\begin{enumerate}
    \item Require aggregating many single-hop answers
    \item Be useful to a wind energy technician
    \item Require exhaustive document search
    \item Create high distractor presence (e.g., for documents from 2018-2022, use  2020-2022 for questions and 2018-2019 as distractors)
\end{enumerate}

If required, we manually correct the resulting pluri-hop question-answer pairs and document citations, ensuring all criteria.

\subsection{Document Analysis}

Distractor density is a key challenge in real corpora based on recurring reports. Distractors are irrelevant passages which are semantically similar to relevant passages (for instance, because they pertain to the wrong entity or time period). For pluri-hop questions requiring the aggregation of data across all documents, such distractors may significantly impair retrieval. Hence, their presence in benchmarks is crucial to generalise the findings to real-world performance. 

To quantify distractor density, we use dataset repetitiveness, i.e. how much text chunks resemble each other in a neighbourhood of size $k$. The repetitiveness at $k$ ($r@k$) is defined as the average cosine similarity between each chunk's embeddings and its $k$ nearest neighbors:

\begin{equation}
r@k = \frac{1}{N} \sum_{i=1}^{N} \frac{1}{k} \sum_{j=1}^{k} \mathrm{cosine\_sim}(x_i, x_{ij}),
\end{equation}

where $x_i$ is chunk $i$ and $x_{ij}$ is its $j$th closest chunk.

For this computation, we randomly sample $N = 100$ documents from PluriHopWIND and 4 other multi-hop datasets (MultiHopRAG \cite{multihoprag}, and the scientific, financial, and legal subsets of Loong \cite{loong}). Each document is chunked into segments of length 500 characters with 100 character overlap and embedded using OpenAI's text-embedding-3-large model. We scan over $k \in \{1, 2, 5, 10, 20, 50\}$.

\begin{figure}
    \centering
    \includegraphics[width=\linewidth]{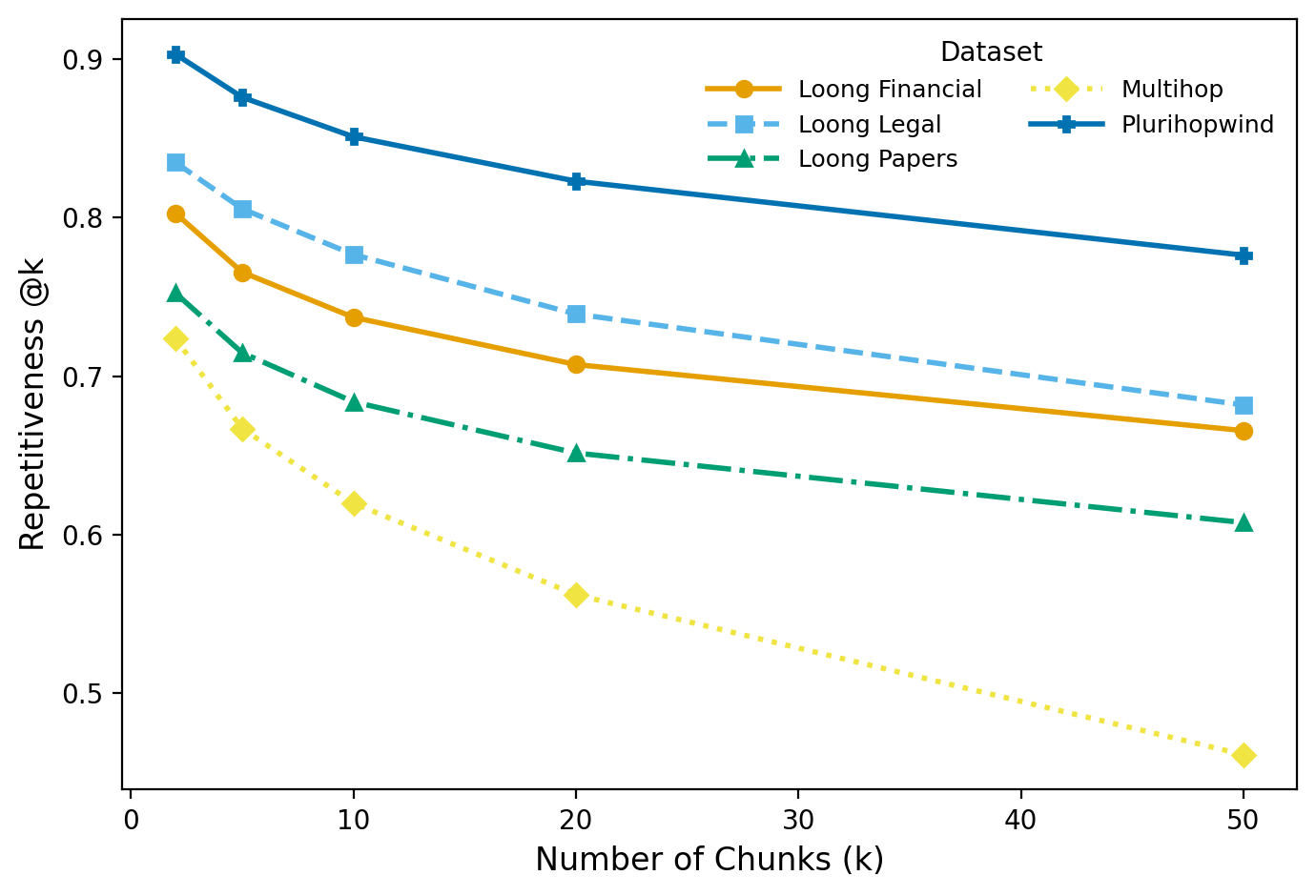}
    \caption{Chunk repetitiveness for PluriHopWIND, Loong, and MultiHopRAG datasets}
    \label{fig:repetitiveness}
\end{figure}

As shown in Figure~\ref{fig:repetitiveness}, PluriHopWIND exhibits 8--20\% higher repetitiveness at $k = 2$ compared to other datasets while at $k = 50$ the relative gap is 13--41\%. Moreover, as $k$ increases PluriHopWIND's repetitiveness drops the least (13\%), indicating that even large top-$k$ retrieval returns highly semantically similar distractors.

These findings show that PluriHopWIND reflects the distractor-rich structure of real recurring report corpora. Naive top-$k$ approaches using similarity based global retrieval struggle in this setting. Instead, approaches that examine documents individually offer a better alternative.

\section{Model}

\subsection{Overview}

\begin{algorithm}
\caption{Model Workflow}
\label{algo:rag_model}
\begin{adjustbox}{max width=\columnwidth,center}
\begin{minipage}{\columnwidth}
\small
\begin{algorithmic}[1]
    \WrapState{\texttt{intermediate\_questions}, \texttt{hypothetical\_summary} $\gets$ DecomposeQuery(\texttt{query})}
    \WrapState{\texttt{candidate\_docs} $\gets$ PerformSimilaritySearchOfSummaries(\texttt{hypothetical\_summary}, \texttt{metadata}, \texttt{K})}
    \State \texttt{intermediate\_answers} $\gets$ [\,]
    \ForAll{\texttt{doc} \textbf{in} \texttt{candidate\_docs}}
        \State \texttt{doc\_chunks} $\gets$ [\,]
        \ForAll{\texttt{question} \textbf{in} \texttt{intermediate\_questions}}
            \WrapState{\texttt{chunks} $\gets$ SimilaritySearchChunks(\texttt{doc}, \texttt{question}, \texttt{k})}
            \State \textbf{append} \texttt{chunks} \textbf{to} \texttt{doc\_chunks}
        \EndFor
        \WrapState{\texttt{relevance} $\gets$ CalculateCrossEncoderScore(\texttt{hypothetical\_summary}, \texttt{doc\_chunks})}
        \If{\texttt{relevance} $> \tau$}
            \ForAll{\texttt{q} \textbf{in} \texttt{intermediate\_questions}}
                \WrapState{\texttt{answer} $\gets$ AnswerIntermediateQuestion(\texttt{q}, \texttt{doc}, \texttt{doc\_chunks})}
                \State \textbf{append} (\texttt{q}, \texttt{answer}) \textbf{to} \texttt{intermediate\_answers}
            \EndFor
        \EndIf
    \EndFor
    \State \texttt{final\_answer} $\gets$ AggregateAnswers(\texttt{intermediate\_answers})
\end{algorithmic}
\end{minipage}
\end{adjustbox}
\end{algorithm}

\begin{figure*}
    \centering
    \includegraphics[width=1\linewidth]{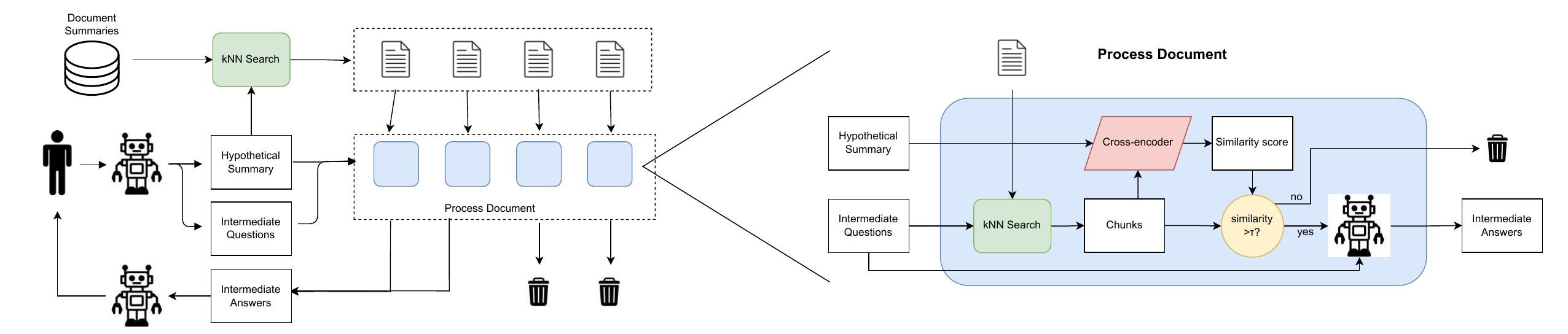}
    \caption{Diagram of PluriHopRAG algorithm}
    \label{fig:plurihoprag-diagram}
\end{figure*}

Our RAG algorithm pseudocode is displayed in Algorithm \ref{algo:rag_model}, and visualized in Figure \ref{fig:plurihoprag-diagram}. There are 3 main differences to a naive RAG pipeline:

\textbf{1. Document-scope-based query decomposition} ($\mathrm{DecomposeQuery}$, line 1): instead of answering a user's question directly, we decompose it into document-scope intermediate questions and then aggregate the document-wise intermediate answers. The decomposition is performed by an LLM fine-tuned with synthetic examples, created from documents and questions about them, see below. In addition to the intermediate questions, $\mathrm{DecomposeQuery}$ also generates a hypothetical summary of a document that would be relevant to answer the original question; this is used for document-wise retrieval. Our query decomposition method is explained further in the next subsection.

\textbf{2. Document filtering using a cross encoder} ($\mathrm{CrossEncoderScore}$, lines 10-11). To minimize LLM token usage for highly exhaustive questions, we estimate each document's relevance to the original question using a cross encoder model, before answering the intermediate questions about that document. We use the cross encoder to calculate the similarity between the hypothetical summary generated by $\mathrm{DecomposeQuery}$, and the concatenation of chunks retrieved for answering all intermediate questions. If the similarity is below a certain threshold, the document is not considered for the question, see below for details.

\textbf{3. Two-step retrieval}. We first retrieve candidate documents by comparing document summaries to a hypothetical relevant document summary from $\mathrm{DecomposeQuery}$ (line 2), then retrieve chunks within each candidate document for each intermediate question (line 7).

\subsection{Query decomposition}

One of the key ingredients of PluriHopRAG is structure-aware query decomposition - rewriting the original query into intermediate questions that reflect how information is organized in the target corpus. This step was motivated by two observations. First, the type of information explicitly requested in a pluri-hop question often differs from each document's content. Second, pluri-hop questions implicitly express filter conditions and aggregation instructions that must be disentangled. For instance, the question "Has Jane Doe's kidney function been steadily declining over the past 3 years?" implicitly contains filter conditions (patient name, time range), a document-level query (kidney function status), and an aggregation instruction (check for decline over time). This can be retrieved by answering these document-level questions:
\begin{enumerate}
    \item Is the person this document talks about Jane Doe?
    \item When was this document written?
    \item What does this document say about the patient's kidney function?
\end{enumerate}

In this scenario, query decomposition becomes unnecessary if there is already a single document containing Jane Doe's 3-year kidney function trend. In other words, the pluri-hop nature of a question is contingent on how evidence is stored in documents which, in turn, the query decomposer needs to understand. This understanding may come from the base LLM's general knowledge imbued during pre-training, but for niche or closed domains it can be introduced through supervised fine-tuning. We propose a workflow where an LLM is fine-tuned for the query decomposition task with fully LLM-generated examples that are created from a subset of the documents from the corpus.

To generate the examples, the LLM is fed tuples consisting of a pluri-hop question and a document relevant in answering it. It is instructed to 

\begin{enumerate}
    \item Reason what information is relevant within the document to answer the question
    \item After the reasoning tokens, generate a list of questions to ask to an equivalent document that would be sufficient to extract all the relevant information
\end{enumerate}

The questions used to create the training set are generated via the same two-step pipeline as the dataset questions - by passing a set of single-document question-answer pairs to an LLM (see Section 3), but without answer verification as we only need the question. We use $N=100$ questions and use OpenAI's supervised finetuning service to fine-tune their GPT-4o model, with $N_{epochs} =3$, learning rate multiplier = 2, and batch size = 1.

\begin{table}[!t]
\centering
\begin{tabular}{lccc}
\toprule
\multicolumn{4}{c}{\textbf{GPT-4o}} \\
\midrule
Setting     & Precision & Recall & F1 \\
\midrule
Fine-tuned  & 0.48 & 0.39 & 0.36 \\
Few-shot    & 0.50 & 0.31 & 0.30 \\
\bottomrule
\end{tabular}
\caption{Comparison of PluriHopRAG performance on PluriHopWIND with GPT-4o as base model, using a fine-tuned vs. few-shot prompted query decomposer}
\label{tab:query-decomposition-ablation}
\end{table}

\subsection{Document filtering}

Given the exhaustive nature of pluri-hop questions, checking all documents with separate LLM calls would be prohibitively expensive. Therefore, we filter out irrelevant documents via cross-encoder filtering based on a commercial pre-trained model (Cohere Rerank 3.5) before LLM reasoning.

Pre-trained reranking models are trained to estimate the relevance of one passage of text to another \cite{rag-review-2023}. Here, the reranking model compares the hypothetical summary and the concatenation of all chunks retrieved to answer the intermediate questions. We discard the entire document if the similarity score output by the reranking model between the hypothetical summary and retrieved chunks is below some threshold ($\tau = 0.1$).

\begin{figure*}
    \centering
    \includegraphics[width=\linewidth]{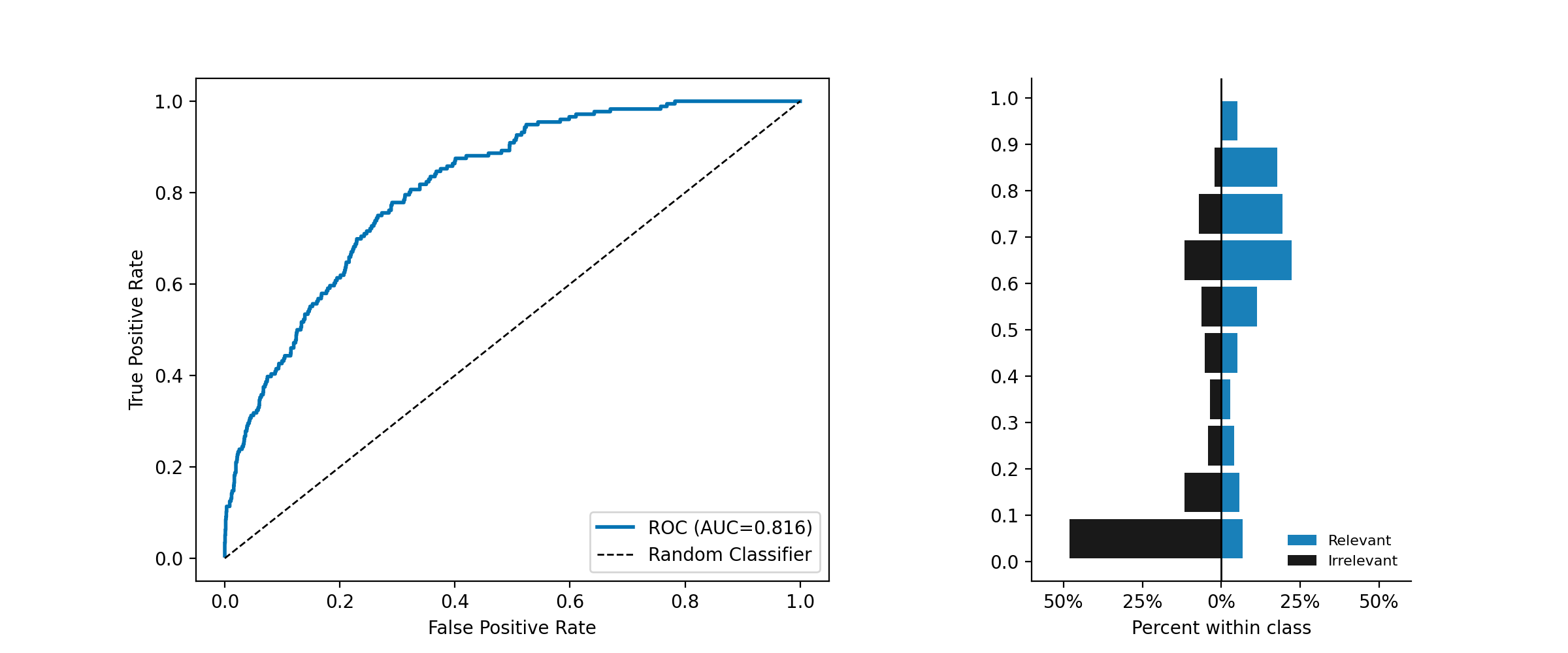}
    \caption{Behavior of cross-encoder based document filter for the PluriHopWIND dataset. Left: Receiver Operator Characteristic (ROC) curve for filter at different filter thresholds $\tau$. Right: distribution of estimated document relevance for relevant and irrelevant documents.}
    \label{fig:filter}
\end{figure*}

Compared to human labels of document relevance to PluriHopWind questions, this approach performs well, see Figure \ref{fig:filter}. The cross-encoder filter correctly removes almost 50\% of the documents manually assessed as irrelevant to each question. On the other hand, only 10\% of relevant documents are removed, showing that the document filter reduces LLM token usage without a great impact on document recall.

\section{Experimental Setup}

\subsection{PluriHopWIND benchmark}

We run our PluriHopWIND benchmark on our PluriHopRAG model, as well as multiple prominent competing RAG approaches: a graph-based RAG \cite{graphrag}, a multi-model RAG \cite{visdomrag}, and a naive RAG baseline \cite{RAG-original-paper}. We test multiple variations of naive RAG, with two chunking methods (standard per-character chunking \& per-page chunking), as well as cross-encoder reranking of chunks.

\textbf{Indexing.} We chunk the document text into chunks with $L=500$ characters each and $l=100$ overlap between them, except for GraphRAG \cite{graphrag} which indexes all data into a knowledge graph.

\textbf{PluriHopRAG.} We retrieve all document summaries ($K > 198$) and set a threshold of $\tau = 0.1$ for the cross-encoder filter.

\textbf{Naive RAG.} We use a basic RAG pipeline \cite{rag-review-2023}, augmented with a pre-trained reranking model (Cohere Rerank 3.5). We put each page's content into one chunk. We also try character-based chunking and no reranking model in the Appendix.

\textbf{GraphRAG and VisdomRAG.} We use the published code to run the models \cite{graphrag, visdomrag}.

\textbf{Evaluation.} We evaluate answers using statement-wise precision, recall, and F1.
For each question, we split the reference answer into a set of atomic statements
$G$ (gold statements), and the model-generated answer into a set of atomic
statements $P$ (predicted statements).

\begin{align}
    \text{Precision} = \frac{|P \cap G|}{|P|}, \\
    \text{Recall} = \frac{|P \cap G|}{|G|}, \\
    \text{F1} = \frac{2\,|P \cap G|}{|P| + |G|}.
\end{align}

The statement-level metrics, inspired by \cite{ragas}, are used instead of more common token-level metrics (such as token-level F1) because they evaluate model outputs at a semantic rather than surface level. This distinction is crucial for PluriHopWIND, where gold standard answers often span multiple sentences and can be expressed through many valid paraphrases.

We split the answers into statements and evaluate the presence of a statement within an answer using GPT-5, with a few-shot prompt that is provided in the Appendix.

\subsection{Loong benchmark}

To demonstrate the general utility of PluriHopRAG, we also evaluate it on Loong \cite{loong}. Loong is built on research papers, legal documents and financial reports. It emphasizes aggregation across many passages of text and high recall sensitivity. However, many questions break the exhaustiveness criterion of pluri-hop (stopping conditions do exist), offering an assessment of PluriHopRAG outside the class of questions it was designed for.

Per domain, we reserve $N=100$ questions for fine-tuning the query decomposer (keeping the same fine-tuning setup as for PluriHopWIND) and evaluate the model on the other $1300$ questions. We use $GPT-4o$, the best performing base model in the original comparison \cite{loong}. The model hyperparameters are the same as for PluriHopWIND. The evaluation scheme is the same as in \cite{loong}:  an LLM judge (GPT-4) assigns a score of 1 to 100 based on accuracy, hallucinations, and completeness.

\section{Results}

\begin{table*}[!t]
\centering
\begin{tabular}{lccc|ccc}
\toprule
\multirow{2}{*}{Method} & \multicolumn{3}{c|}{Claude 4 Sonnet} & \multicolumn{3}{c}{GPT-4o} \\
 & Precision & Recall & F1 & Precision & Recall & F1 \\
\midrule
PluriHopRAG & \textbf{0.47} & \textbf{0.57} & \textbf{0.44} & 0.48 & \textbf{0.39} & \textbf{0.36} \\
NaiveRAG & 0.47 & 0.43 & 0.38 & \textbf{0.64} & 0.26 & 0.25 \\
VisdomRAG & 0.39 & 0.12 & 0.19 & 0.32 & 0.24 & 0.21 \\
GraphRAG & 0.34 & 0.36 & 0.30 & 0.40 & 0.22 & 0.21 \\
\bottomrule
\end{tabular}
\caption{QA performance on PluriHopWIND. For NaiveRAG, the best configuration per base LLM is reported (see Table~\ref{tab:naiverag_ablation} for details).}
\label{tab:model_comparison_simplified}
\end{table*}

\begin{table*}[!t]
\centering
\begin{tabular}{llccccc}
\toprule
Method & Base LLM & Set 1 & Set 2 & Set 3 & Set 4 & Overall \\
\midrule
\multicolumn{2}{l} --
& 10K--50K & 50K--100K & 100K--200K & 200K--250K & -- \\
\midrule
PluriHopRAG & GPT-4o 
& \textbf{81.75} & \textbf{75.20} & \textbf{68.23} & \textbf{56.48} & \textbf{68.53} \\
Long-context & GPT-4o 
& 70.40 & 58.38 & 46.95 & 31.11 & 51.71 \\
NaiveRAG & GPT-4o 
& 50.55 & 49.96 & 45.99 & 33.82 & 45.08 \\
\midrule
StructRAG & Qwen2-72B 
& 69.43 & 60.95 & 57.92 & 51.42 & 59.93 \\
RQ-RAG & Qwen2-72B 
& 53.51 & 47.09 & 40.93 & 31.91 & 43.36 \\
GraphRAG & Qwen2-72B 
& 40.82 & 33.06 & 33.28 & 23.47 & 32.66 \\
\bottomrule
\end{tabular}
\caption{QA performance on the Loong benchmark \cite{loong}. StructRAG, RQ-RAG, and GraphRAG results are taken from \cite{structrag}, long-context and NaiveRAG results are taken from \cite{loong}.}
\label{tab:loong_setwise_comparison}
\end{table*}

The performance results of RAG models on PluriHopWIND and Loong are shown in tables \ref{tab:model_comparison_simplified} and \ref{tab:loong_setwise_comparison}, respectively. The main conclusions are as follows:

\textbf{PluriHopRAG achieves a significantly higher answer F1 score than other tested models across base LLMs.} We see a 18\% relative improvement (0.4 to 0.47) in F1 score with Claude 4 Sonnet as the base model and a 52\% relative improvement with GPT-4o (0.27 to 0.41). In both cases the second best model is naive RAG with Cohere Rerank 3.5 as reranker, significantly outperforming naive RAG without a reranking model.

\textbf{PluriHopWIND offers a very difficult challenge for modern RAG systems.} Despite its modest size, PluriHopWIND exposes fundamental weaknesses in modern QA systems when aggregating data in an exhaustive fashion from repetitive, distractor-rich report corpora - a set of requirements that is common in manufacturing, medicine, finance and other fields.

\textbf{PluriHopRAG generalizes well to other domains, showing performance gains on Loong.} We report a 33\% relative increase in performance over long-context reasoning (passing full documents to LLM), a 52\% increase over Naive RAG, and 14\% increase over the previously best performing RAG model (StructRAG). We note that StructRAG was only evaluated on a different base LLM, but due to the expensive 
reinforcement learning procedure required to train the model we deemed evaluating it on GPT-4o prohibitively expensive.

\subsection{Ablations}

\subsubsection{Fine-tuned query decomposition}

We evaluate our model on PluriHopWIND with a fine-tuned query decomposer and one based on the base version of GPT-4o, only adding a few-shot prompt with $N=2$ examples from the training set of the query decomposer. The results are in Table \ref{tab:query-decomposition-ablation} - fine-tuning adds a 20\% relative increase in F1 score. Comparing the few-shot version to other models from Table \ref{tab:model_comparison_simplified}, it is evident that fine-tuning is crucial to achieve noticeable performance increases over baseline models. This adds weight to our claim that the basic logic of how information is laid out in document corpora can be imbued using fine-tuning with very modest training set sizes.

\section{Conclusion}

In this work, we formalized the notion of pluri-hop questions - queries that possess both high recall sensitivity, exhaustiveness (no clear stopping condition), and exactness (factual questions with an unambiguously best answer). Such questions arise naturally in domains with recurring report data but they are poorly represented in existing benchmarks.

To study this challenge, we developed PluriHopWIND, a diagnostic dataset constructed from real wind industry technical reports. Its design emphasizes distractor-heavy, repetitive corpora that cannot fit within an LLM’s context window, thereby replicating the practical difficulties of answering pluri-hop questions. Using our proposed intersimilarity measure of distractor density, we showed that PluriHopWIND more closely resembles realistic pluri-hop scenarios than comparable benchmarks. By focusing on distractor density during dataset construction, we managed to showcase failure modes of RAG systems in realistic scenarios despite its modest size.

We also presented PluriHopRAG, a retrieval architecture tailored to the pluri-hop setting. Its core insight is that effective query decomposition must reflect corpus-specific document structure—knowledge that can be learned from a few synthetic examples without manual annotation. Combined with cross-encoder filtering for exhaustive but cheap document coverage, PluriHopRAG achieves relative F1 gains of 18-52\% on PluriHopWIND depending on the base LLM, and generalizes to financial, legal, and research domains with notable gains on the Loong benchmark.

Together, these contributions extend the RAG literature in three directions: (1) formalizing pluri-hop questions as a distinct category from traditional multi-hop reasoning, (2) providing a dataset exemplifying the challenges of real-world recurring report corpora, and (3) demonstrating that learning document structure enables more effective retrieval for exhaustive, recall-sensitive question answering.

\section{Limitations}

While our study introduces new concepts and methods for pluri-hop QA, it also comes with limitations:

\textbf{Dataset size and coverage.}
PluriHopWIND contains 191 documents and 48 questions, which is modest compared to other QA benchmarks \cite{2wikimultihopqa, hotpotqa, monaco, loong, mebench}. Given the high repetitiveness of the dataset, we believe this size is sufficient for a showcase of the difficulty of pluri-hop QA and the shortcomings of top-k retrieval for this question type. Nevertheless, broader validation across larger and more diverse corpora is necessary to advance in this space.

\textbf{Incomplete comparison to StructRAG}
The best-performing model on Loong to date has been StructRAG \cite{structrag}, which has only been evaluated on Loong with Qwen 2-72b as the base LLM. We focussed on evaluating PluriHopRAG on the best performing base LLM among those tried in \cite{loong}. Due to the Direct Preference Optimization procedure used in StructRAG, we deemed training the model on GPT-4o for Loong too expensive.


\section*{Acknowledgements}

We gratefully acknowledge the financial support provided by the State of Berlin within the framework of the program "Programm zur Förderung von Forschung, Innovationen und Technologien – Pro FIT." This work was conducted as part of the project "Copilot – Entwicklung eines Chatassistenten für Betreiber von Windenergieparks." We thank Carlotta Fabian for contributing to the initial development of ideas for PluriHopRAG and for their work on the code development and review. We thank Alessandra Cossu, Clemens Edler and Carlotta Fabian for taking part in the manual verification of PluriHopWIND. We thank Aida Takhmazova, Carlotta Fabian, Mateusz Majchrzak and Michael Tegtmeier for feedback on an earlier version of the manuscript. We thank Agnieszka Michalska for helping create the figures for the manuscript.


\bibliography{bibliography}

\appendix


\section{LLM Prompts}

\subsection{Question Decomposition}

\begin{promptbox}[title=Prompt: Question Decomposer (Base Version)]
I have a RAG application.  
Given a question about one or multiple documents, determine:

1. A hypothetical summary of the document (or one of the documents) that would be relevant to answer the question (max 100 tokens).  
2. A set of questions to ask to the document(s) to retrieve all information needed to answer the question.

\vspace{4pt}
\textbf{Rules:}
\begin{itemize}
    \item Sometimes multiple documents are needed to answer the question. So a question about a trend could be answered either with a document describing this trend (if such a document exists, usually it doesn't), or with multiple documents describing the current situation and the trend could be inferred. Therefore, the questions should take both possibilities into account.
    \item Try to get all needed information with as few questions as possible, minimizing overlap.
\end{itemize}

\vspace{4pt}
Return in JSON format, without markdown code block formatting, as follows:  
\{\{'hypothetical\_summary': str, 'questions': list[str]\}\}
\end{promptbox}

\subsection{Document-level Answering}

\begin{promptbox}[title=Prompt: Document Answer Generator]
You are a wind energy expert.  
Given one or multiple questions, answer all of them using the provided context.  
All the context comes from one document.

Return in JSON format, without markdown code block formatting,  
with key \texttt{'answers'} and value list of strings.

\vspace{4pt}
\textbf{Inputs:}  
Questions: \{questions\}  
Context: \{context\}
\end{promptbox}

\begin{promptbox}[title=Prompt: Page Group Answer Aggregator]
I tried to answer multiple questions using individual pages or groups of pages from a document.  
Given the answers based on each page, construct the correct answers based on the whole document.  

Return in JSON format, without markdown code block formatting,  
with key \texttt{'answers'} and value a list of strings.  

Do not omit any relevant details.

\vspace{4pt}
\textbf{Inputs:}  
Questions: \{questions\}  
Answers: \{answers\}
\end{promptbox}

\subsection{Corpus-level Aggregation}

\begin{promptbox}[title=Prompt: Answer Aggregator]
A question was asked about some document(s).  
This question was split into intermediate questions, and these intermediate questions were answered with one or multiple documents as context.  

Given the original question, the intermediate questions, and each document's answer to the intermediate questions, construct the final answer to the original question (in the language of the original question).  

Only include information that directly answers the original question.  
If that means omitting some information from the intermediate answers, that's fine.  
Don't explain how you arrived at the answer.

After each fact, put a reference to the document with \texttt{[Document <document\_index>]}.  
If a fact comes from multiple documents, reference them like \texttt{[Document <1>]}, \texttt{[Document <2>]}, etc., instead of \texttt{[Document 1, 2]}.

After you construct the final answer, also return a list of documents which were relevant to answer the question  
(i.e. all documents you referenced, in ascending order of index).  

The output should be in JSON format.  

\vspace{4pt}
\textbf{Example Output:}  
\{\{'answer': 'example answer', 'relevant\_documents': [3, 5, 6]\}\}

\vspace{4pt}
\textbf{Your Task:}  

Original Question: \{original\_question\}  
Intermediate Questions: \{intermediate\_questions\}  
Document Answers: \{document\_answers\}  

Final Answer (RETURN IN JSON, without markdown code block formatting):
\end{promptbox}

\subsection{Evaluation - Statement Splitting}

\begin{promptbox}[title=Prompt: Statement Splitter (for Answers)]
Below is a question and answer. I want to split the answer into statements in such a way, that I can recreate the answer (or a paraphrased version) by using the question and the statements, while keeping the statements as few and as short \& simple as possible. If it makes sense, the statements should be key-value pairs (with keys and values as strings), otherwise they should be strings. The whole answer should be in json format, in the following format: \\
\{ \\
\quad "1": <statement\_1>, \\
\quad "2": <statement\_2>, \\
\quad ... \\
\}

Below are some rules to follow:
1. There should be as few statements as possible, and they should be as simple as possible, to still recreate the answer (or a paraphrased version of the answer) using BOTH the question and statements. \\
Example: \\
Question: \\
Are there any anomalies in the oil report for wind turbine 123? \\
Answer: \\
Yes there are 2 anomalies in the oil report for wind turbine 123: the chrome level is too high and the magnesium level too high. \\
Bad outcome: \\
\{ \\
\quad "1": \{"turbine": "123"\} \# this statement isn't necessary to recreate the answer because the turbine id can be found in the question \\
\quad "2": \{"number of anomalies in oil report": "2"\} \# it's unnecessary to write "oil report" because the document type can be found in the question \\
\quad "3": \{"anomaly": "chrome level too high"\} \\
\quad "4": \{"anomaly": "magnesium level too high"\} \\
\} \\
Desired outcome: \\
\{ \\
\quad "1": \{"number of anomalies": "2"\}, \\
\quad "2": \{"anomaly": "chrome level too high"\}, \\
\quad "3": \{"anomaly": "magnesium level too high"\} \\
\}

2. If the statement is a string, it should be max 1 short sentence. If it is a key-value pair, the value must be max 1 short sentence. \\
Example: \\
"Conclusion: Chromium levels high. Continue monitoring to observe further trends" \\
Desired behaviour: \\
\{ \\
\quad "1": \{"Conclusion": "Chromium levels high"\}, \\
\quad "2": \{"Conclusion": "Continue monitoring to observe further trends"\} \\
\}

3. If an answer is refused because relevant context couldn't be found, and alternative questions are suggested to avoid this, this should be interpreted as zero statements. If the answer is that relevant context couldn't be found, but the irrelevant context is talked about anyway, the answer should be treated like any other.

4. If the answer contains references to documents via their filenames, this should be ignored and not included in the inferred statements.

Question: \\
\{question\}

Answer: \\
\{answer\}
\end{promptbox}

\subsection{Evaluation - Statement Comparison and Counting}

\begin{promptbox}[title=Prompt: Statement Counter and Comparator]
An answer to a question was split into statements. You need to compare this answer to another, reference, answer. For each statement, determine SEPARATELY if the *exact* statement can be directly implied from the reference answer (not the original answer)?. Respond in json format, where for each statement the key is the statement index and the value is a bool that is true if you can infer the statement from the text, false otherwise. Also have a key-value pair where the key is "inferred\_statements" and the value is the number of keys in the dictionary with value true.

EXAMPLE:
Answer: In the past 5 years, the repairs on wind turbine 123 have occured 4 times: on 2020.05.01, 2021.05.02, 2022.05.04, and 2023.05.04. \\
Statements: ['\{\{number of repairs': '4'\}\}', '\{\{repair date': '2020.05.01'\}\}', '\{\{repair date': '2021.05.02\}\}', '\{\{repair date': '2022.05.04\}\}', '\{\{repair date': '2023.05.04\}\}'] \\
Reference text: There were 5 repairs conducted in the past 5 years: on 2020.05.01, 2021.05.02, 2022.05.03, 2023.05.04, and 2024.05.05. \\
EXAMPLE OUTPUT: \{'1': false, '2': true, '3': true, '4': false, '5': true, 'inferred\_statements': 3\}

YOUR TASK: \\
Answer: \{text\} \\
Statements: \{statements\} \\
Reference text: \{reference\_text\}
\end{promptbox}

\begin{figure*}[tbp]
    \includegraphics[width=0.95\textwidth]{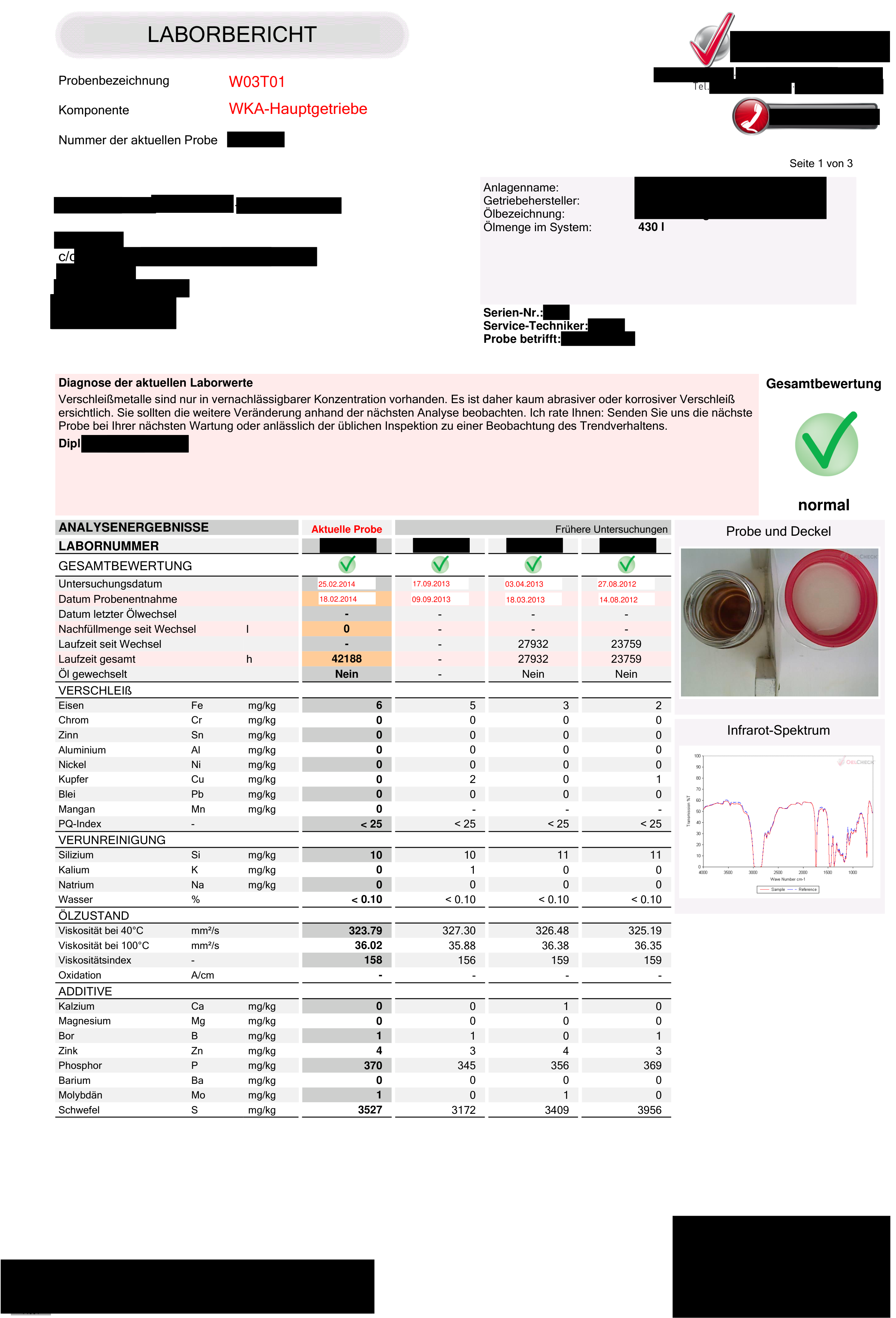}
    \caption{Typical report in the PluriHopWIND dataset}
    \label{fig:typical_report}
\end{figure*}

\subsection{Naive RAG hyperparameter comparison}

Here we evaluate different variants of Naive RAG on PluriHopWIND. We compare the performance of Naive RAG for two different chunking methods - one chunk per page, and chunks of equal size ($L=500$, $l=100$) - and with/without reranking model (selecting $k'=20$ chunks out of $k=80$). The results are in Table \ref{tab:naiverag_ablation} - per-page chunking with reranking performs best, and is thus selected for the RAG model comparison.

\begin{table*}[!t]
\centering
\begin{tabular}{lccc|ccc}
\toprule
\multirow{2}{*}{NaiveRAG Variant} & \multicolumn{3}{c|}{Claude 4 Sonnet} & \multicolumn{3}{c}{GPT-4o} \\
 & Precision & Recall & F1 & Precision & Recall & F1 \\
\midrule
Per-page chunking & \textbf{0.50} & 0.30 & 0.26 & 0.75 & 0.14 & 0.14 \\
Per-page chunking + rerank & 0.48 & \textbf{0.47} & \textbf{0.40} & 0.62 & \textbf{0.26} & \textbf{0.27} \\
Char-count chunking & 0.47 & 0.18 & 0.17 & \textbf{0.81} & 0.10 & 0.12 \\
Char-count chunking + rerank & 0.44 & 0.36 & 0.31 & 0.65 & 0.21 & 0.21 \\
\bottomrule
\end{tabular}
\caption{NaiveRAG ablation over chunking strategy and reranking. Reranking consistently improves recall and F1 across base LLMs, while optimal chunking differs by model.}
\label{tab:naiverag_ablation}
\end{table*}

\end{document}